\documentclass[10pt, a4paper]{article}
\usepackage{threeparttable}
\usepackage{booktabs}
\usepackage{tabularx}
\usepackage{pifont}
\usepackage{subcaption}
\usepackage{multirow}
\usepackage{graphicx}
\usepackage{makecell}
\usepackage[final]{lrec2026} 

\title{S-VoCAL: A Dataset and Evaluation Framework for Inferring Speaking Voice Character Attributes in Literature}

\name{Abigail Berthe-Pardo$^{1}$, Gaspard Michel$^{1,2}$, Elena V. Epure$^{2,3}$, Christophe Cerisara$^{1}$} 

\address{
$^{1}$LORIA, Vandœuvre-lès-Nancy, France \\
$^{2}$Deezer Research, Paris, France \\
$^{3}$Idiap Research Institute, Switzerland \\
abigail.berthe@hotmail.fr, gmichel@deezer.com, elena.epure@idiap.ch , christophe.cerisara@loria.fr
}

\abstract{With recent advances in Text-to-Speech (TTS) systems, synthetic audiobook narration has seen increased interest, reaching unprecedented levels of naturalness.
However, larger gaps remain in synthetic narration systems' ability to impersonate fictional characters, and convey complex emotions or prosody.
A promising direction to enhance character identification is the assignment of plausible voices to each fictional characters in a book.
This step typically requires complex inference of attributes in book-length contexts, such as a character's age, gender, origin or physical health, which in turns requires dedicated benchmark datasets to evaluate extraction systems' performances.
We present S-VoCAL (Speaking Voice Character Attributes in Literature), the first dataset and evaluation framework dedicated to evaluate the inference of voice-related fictional character attributes.
S-VoCAL entails 8 attributes grounded in sociophonetic studies, and 952 character-book pairs derived from Project Gutenberg.
Its evaluation framework addresses the particularities of each attribute, and includes a novel similarity metric based on recent Large Language Models embeddings.
We demonstrate the applicability of S-VoCAL by applying a simple Retrieval-Augmented Generation (RAG) pipeline to the task of inferring character attributes.
Our results suggest that the RAG pipeline reliably infers attributes such as Age or Gender, but struggles on others such as Origin or Physical Health.
The dataset and evaluation code are available at \href{https://github.com/AbigailBerthe/S-VoCAL}{https://github.com/AbigailBerthe/S-VoCAL}.
\newline 
}

\begin{document}

\maketitleabstract

\section{Introduction}
Audiobooks have recently attracted growing worldwide interest:
the Audiobook industry in the United-States, Europe and China revenue is projected to increase by 20\% by the end of 2025 \cite{statistaChina, statistaEurope, statistaUSA}.
Among reasons for success, the ability of Audiobooks to promote literature to individuals that usually do not or cannot engage with the act of reading, such as individuals with reading or visual impairments along with the ability to perform other actions while listening is often mentioned \cite{roy2024audiobooks, impairementscitation}.
Meanwhile, advances in text-to-speech (TTS) systems' ability to synthesize natural and expressive speech have sparked interest in the synthetic reading of Audiobooks \cite{walsh2023largescaleautomaticaudiobookcreation, park2025multiactoraudiobookzeroshotaudiobookgeneration}, allowing consumers to listen to personalized artificial narrators reading their favorite e-book, and authors to independently promote their work with lower costs than traditional Audiobook production.



A key aspect in performative audiobook narration is the narrator's ability to impersonate fictional characters \cite{brownaudiobooks, goodwinaudiobooks}.
For example, narrators typically employ lower pitch when reading utterances from male character or from narrative parts \cite{Pethe_2025}.
For listeners, this aspect of narration is essential to enhance character identification and reading comprehension \cite{van2017evoking, wolters2022reading} and has been identified as a critical limitation of current synthetic Audiobooks \cite{SyntheticReading2023}.
Besides, modern TTS systems typically suffer from lack of expressivity when synthesizing fictional character utterances \cite{michel2025libriquotespeechdatasetfictional}, positioning the enhancement of character impersonation in artificial readers as a central objective for achieving broader advancements in Audiobook narration synthesis.

The alignment between plausible voices and fictional character descriptions yields a promising direction that has been recently explored in the TTS community~\cite{xin2022improvingspeechprosodyaudiobook, park2025multiactoraudiobookzeroshotaudiobookgeneration}.
However, finding these character-appropriate voices requires complex Natural Language Processing (NLP) steps to extract voice-specific attributes such as age, gender, region of origin, or physical condition for each character.
This information might not always be explicitly stated in books and might be disseminated in distant narrative sections, making the precise extraction of such attributes challenging \cite{iosif-mishra-2014-speaker, papoudakis-etal-2024-bookworm}.
To evaluate the ability of NLP systems to accurately extract character information, prior dataset efforts have mostly proposed general descriptions of characters in natural language drawn from literature analysis websites \cite{yuan-etal-2024-evaluating, Gurung_2024, papoudakis-etal-2024-bookworm}.
Such descriptions are usually centered around structured profiles or complex literary analysis, involve various side-information that do not correlate with a character's voice or might critically miss important voice-related attributes.

Hence, we propose S-VoCAL (Speaking Voice Character Attributes in Literature), a dataset and evaluation framework dedicated to the extraction of fictional character attributes that correlate with speaking voice delivery.
S-VoCAL entails 8 attributes grounded in phonetic studies, and 952 characters-book pairs distributed among 192 unique novels collected from Project Gutenberg\footnote{\url{https://www.gutenberg.org/}}.

Attributes are initially extracted by matching characters with their Wikidata\footnote{\url{https://www.wikidata.org/}} entry, from which the \textit{Age} attribute was complemented with manual annotations.
Given the unique and unbalanced nature of each attributes (\textit{e.g.} \emph{Gender} is categorical ; \emph{Physical Health} is a short text ; the \emph{Type} attribute mostly contains \emph{Humans}), we design a dedicated evaluation framework including weighted metrics, and a new semantic similarity metric based on Qwen-3 8b embeddings \cite{zhang2025qwen3embeddingadvancingtext} that we show correlates strongly with human judgments.

Following \citet{papoudakis-etal-2024-bookworm}, we evaluate a simple Retrieval-Augmented-Generation (RAG) approach on the extraction of S-VoCAL character attributes.
The solution involves a Passage Retrieving step based on embeddings derived from instruction-based encoder and an Attribute Inference step performed by Large Language Models (LLMs).
We show that the proposed RAG approach typically performs well at extracting closed-class attributes such as \textit{Age}, \textit{Gender} or \textit{Spoken Languages}, but struggles for open-class attributes such as \textit{Physical Health} and \textit{Occupation}.

To summarize, our contributions are as follows:
\begin{itemize}
    \item We propose S-VocAL, a dataset and evaluation framework for inferring speaker voice character attributes in novels. The data is publicly available and provides 8 attributes for 952 character-book pairs, over 192 unique books.
    \item We design an evaluation framework that accounts for the unique and unbalanced nature of each attributes.
    \item We evaluate a RAG solution for the Attribute Inference task, showing that while it performs well for closed-class attributes, it typically struggle when inferring open-class attributes.
\end{itemize}

\section{Related Works}

\begin{table*}[!ht]
\centering
\small
\setlength{\tabcolsep}{5pt} 
\renewcommand{\arraystretch}{1.3} 
\begin{tabularx}{\textwidth}{
    >{\raggedright\arraybackslash}p{1.5cm} 
    >{\raggedright\arraybackslash}p{3.5cm} 
    >{\raggedright\arraybackslash}p{4cm}
    >{\centering\arraybackslash}p{2.5cm} 
    >{\centering\arraybackslash}p{1.8cm} 
    >{\centering\arraybackslash}p{1cm}
}
\toprule
\textbf{Attribute} & \textbf{Voice relevance} & \textbf{Wikidata properties} & \textbf{Type} & \textbf{Coverage} & \textbf{Is New}\\
\midrule
Name &  & \textit{label} & Text & 100\% & \ding{55}\\
Aliases &  & \textit{also known as, given name, native label, nickname, pseudonym} & List & 73.5\% & \ding{55}\\
Age$^*$ & pitch range, formant frequencies, speech rate & \textit{relative age} & Categorical (4) & 4.1\%$\;|\;$37.7\% & \ding{55}\\
Gender & pitch range, resonance & \textit{sex or gender} & Categorical (2) & 96.5\% & \ding{55}\\
Origin & accent, phonetic features, prosody & \textit{country of citizenship, place of birth, country of origin} & List & 58.8\% & \ding{51}\\
Residence & accent, speech patterns & \textit{residence, work location, country, place of residence} & List & 26.5\% & \ding{51} \\
Occupation & characteristic speech styles & \textit{occupation, character type, position held, noble title, military/police rank, crew membership} & List & 57.7\% & \ding{51} \\
Spoken languages & pronunciation, intonation & \textit{languages spoken, native language} & List & 49.2\% & \ding{51} \\
Physical health & specific voice qualities (e.g., breathiness) & \textit{medical condition} & Text & 4.1\% & \ding{51} \\
Type & human/non-human voice & \textit{instance of} & Text & 100\% & \ding{51} \\
\bottomrule
\end{tabularx}

\caption{Attributes overview in S-VoCAL, along with their mapped Wikidata properties. Age$^*$ is represented as a relative category inferred from voice features, with coverage reported as (initial | final, after manual annotation). The \textit{Is New} column shows originality compared to \cite{iosif-mishra-2014-speaker}.}
\label{gold_ds}
\end{table*}

\subsection{Voice Characteristics}

Voices can be described through a combination of segmental and suprasegmental features \cite{Nolan_1987, di2013prosodie}.
The latter correspond to acoustic features such as pitch, loudness, rhythm, that can be rendered through various lexical items in literary texts \cite{DBU_PARRE_2002_01_0039}.
Besides, voices also convey indexical cues that can be used to infer characteristics of the speaker \cite{laver1979phonetic, abercrombie1967elements, 10.1093/oxfordhb/9780198743187.001.0001}.
We find, among features that have been well studied, the speaker gender and age \cite{Nolan_1987, doi:10.1044/jshr.1501.155, Sorokowski2024VoicebasedJO}, physical health \cite{doi:10.1044/jshd.2803.221} or region of origin and social class \cite{labov1973sociolinguistic}.

These prior works allow us to identify 8 salient acoustic features that correlate with voice characteristics and that are likely to be identifiable from character descriptions in novels. We describe these attributes in Section~\ref{sec:dataset}.

\subsection{Character Attributes Datasets}

Extracting reliable descriptions of characters from narrative texts is a complex task, as relevant information may be implicit, evolve over time, or distributed over several chapters.
Prior works have targeted specific character aspects, such as social roles \cite{stammbach2022heroesvillainsvictimsgpt3, gervas2024tagging}, latent personas \cite{bamman2014bayesian}, and more recently, natural language descriptions \cite{brahman2021let, yuan-etal-2024-evaluating,Gurung_2024,papoudakis-etal-2024-bookworm}.
Such descriptions are either general summary of characters, literary analysis, or centered around targeted dimensions (personality, events, \dots).
In contrast, we define a discrete set of character attributes that correlate with voice delivery, and aim at extracting exactly these attributes rather than high-level character descriptions.

Several works also aim at extracting discrete set of static attributes, such as gender or age \cite{iosif-mishra-2014-speaker, jaipersaud2024showdonttelluncovering, 10447353}, or dynamic attributes such as emotion arcs \cite{vishnubhotla-etal-2024-emotion}.
Similar to S-VoCAL, \citet{iosif-mishra-2014-speaker} targets voice-related attributes for storytelling applications, and propose a pipeline to automatically extract these attributes.
S-VoCAL extends this approach by covering novels rather than children stories, and by entailing a broader set of attributes derived from sociophonetic studies.

\subsection{Evaluating Attribute Inference}

Evaluating a system's ability to infer character attributes is particularly challenging given the unique and unbalanced nature of attributes such as age or physical health.
When attributes are described in natural language, such as in \citet{papoudakis-etal-2024-bookworm}, the evaluation becomes even more challenging, as it requires additional reasoning to ensure factual overlaps between the predicted and ground-truth descriptions.
Therefore, evaluations often include fact-based metrics between predicted and reference descriptions, derived from Large Language Models (LLMs) judgments such as PRISMA \cite{mahon2024modularapproachmultimodalsummarization} or Consistency Score \cite{yuan-etal-2024-evaluating}.
For discrete set of attributes, evaluation approaches also include LLM judgments \cite{jaipersaud2024showdonttelluncovering}, human evaluations \cite{10447353}, and standard metrics such as F1-score \cite{iosif-mishra-2014-speaker}.

Given the unique nature of the selected attributes, we also adopt standard strict and soft F1-metrics for closed and semi-closed class attributes (attributes that have a finite set of possible values).
For open-class text attributes, we leverage instruction-aware LLM embeddings with Qwen-3 8b \cite{zhang2025qwen3embeddingadvancingtext} to calculate semantic similarities between the predicted and ground-truth attribute, similar to BERTScore \cite{bertscore}.
Instruction-aware embedding models allows us to derive attribute-specific representations, which can adapt to each attribute particularity.
We show that the proposed similarity metric correlates better with human judgments than BERTScore.


\section{The Attribute Inference Task}


We define a taxonomy of attributes relevant to vocal identity that can be inferred from text, and used to design a general synthetic voice for a fictional character. We target stable traits, related to the character's identity, rather than low-level acoustic properties.
Voices carry indexical information that enable listeners to infer relatively stable traits, such as age, gender, physical condition, or regional origin \citep{abercrombie1967elements, laver1991phonetic}.
According to \citet{DBU_PARRE_2002_01_0039}, these cues reflect both \textit{intrinsic} aspects, linked to physiological features, and \textit{extrinsic} aspects, shaped by social and contextual factors.
In the context of synthetic book reading, both dimensions are crucial to adapt acoustic parameters of the generated voice, such as pitch range or prosody, to the persona of a fictional character. 

These links between a person's traits and the acoustic realization of their voice have been extensively documented in phonetic and sociophonetic studies. For example, pitch range and fundamental frequency are known to vary with age and gender: male speakers present lower pitch range than females, which in turn exhibit lower pitch range than children \citep{cruttenden1986intonation}, and average pitch ranges tends to decrease with age \citep{hollien1972speaking}. Health can also impact voice quality and pitch stability \citep{canter1963speech}, while geographical origin is shown to correlate with segmental realizations (i.e. pronunciation of individual sounds), and accents \citep{labov1973sociolinguistic}. 

We identify eight voice-related attributes that can be inferred from narrative text: \textit{Age}, \textit{Gender}, \textit{Origin}, \textit{Residence}, \textit{Spoken languages}, \textit{Occupation}, \textit{Physical health}, and \textit{Type}. Their specific influence on vocal characteristics is detailed in Table~\ref{gold_ds}.  

The Attribute Inference task aims at extracting these voice-related attributes from long narrative texts: given a character and a novel, we want to infer traits that can influence how that character’s voice would sound. In practice, this corresponds to predicting one or several values for each of the voice-relevant attribute based on the book content only.
This task is particularly challenging as literary texts seldom state such information explicitly. Relevant cues are often implicit and distributed across distant passages, requiring reasoning over long context to infer certain attributes indirectly.
Accurate reasoning over very long contexts is typically an area where recent NLP models struggle, adding an additional complexity layer to the task of Attribute Inference \cite{hsieh2024rulerwhatsrealcontext, wang2024adalevalevaluatinglongcontextllms}

To the best of our knowledge, no previous dataset provides explicit labels for speaking voice attributes, nor general evaluation frameworks for evaluating Attribute Inference.
Thus, we propose S-VoCAL dataset, a dataset dedicated to the Attribute Inference task, described further.

\begin{figure}
    \centering
    \begin{subfigure}[t]{0.42\linewidth}
\includegraphics[width=\linewidth]{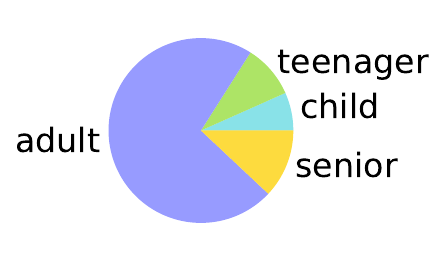}
    \caption{Age}
    \end{subfigure}
    \hfill
    \begin{subfigure}[t]{0.42\linewidth}
\includegraphics[width=0.74\linewidth]{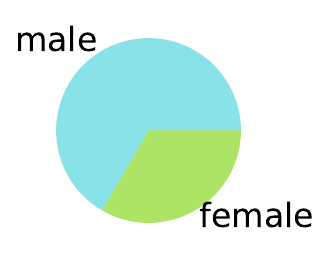}
    \caption{Gender}
    \end{subfigure}
    \vfill
    \begin{subfigure}[t]{0.5\linewidth}\includegraphics[width=\linewidth]{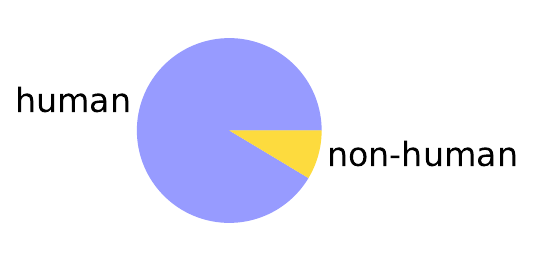}
    \caption{Type}
    \end{subfigure}
    \caption{As attributes from S-VoCAL comes from various Gutenberg books, they are unbalanced, especially for \textit{Type} that mostly contains humans and \textit{Age} that mostly contains adults.}
    \label{fig:distributions}
\end{figure}

\section{S-VoCAL Dataset}
\label{sec:dataset}
S-VoCAL (Speaking-Voice Character Attributes in Literature) focuses on 8 voice-related attributes defined in Table~\ref{gold_ds}, for 952 characters from 192 novels.
These books correspond to works of English or translated world literature published before 1940 and available in full text on Project Gutenberg\footnote{https://www.gutenberg.org/}.
The final selection includes all books that (1) had an English full-text version on Project Gutenberg, (2) had a corresponding entry on Wikidata\footnote{https://www.wikidata.org}, and (3) included at least one character on Wikidata (main or secondary), with any of the target properties used to create our attributes (see Table \ref{gold_ds}).
Figure~\ref{fig:distributions} presents the distribution of characters across the main categorical attributes (Age, Gender, and Type), with "adult", "male", and "human" characters largely predominant in the corpus.

\subsection{Data Collection}

Prior works have mostly relied on character descriptions from literary summary websites or fan-wikis \cite{papoudakis-etal-2024-bookworm, Gurung_2024, yuan-etal-2024-evaluating}.
However, these sources vary considerably in the level of detail across books and characters, and exploiting them would have required an additional extraction pipeline with its own validation process.
We therefore opted for Wikidata as our main source of character information.
Wikidata is a collaborative knowledge graph that provides structured properties about entities, including literary works and fictional characters, filled in by human contributors.
Its structured format and frequent human activity make it particularly reliable to align existing properties with our target attributes.

We organized the dataset based on the available characters' properties in Wikidata. For each character, these properties were heterogeneous and mostly did not match directly our target attributes. We manually selected properties that were semantically very close to the target attribute (either synonyms or narrower subsets) and grouped them under the corresponding predefined attributes. In some cases, multiple complementary values were merged into the same attribute. E.g. \textit{Spoken Languages} aggregates several Wikidata properties, such as \textit{languages spoken}, \textit{native language}, \textit{language from name}, and \textit{writing language}, while the attribute \textit{Gender} is directly mapped to a single Wikidata property, \textit{sex or gender}. Table~\ref{gold_ds} details the mapping of Wikidata property per attribute.

After collection, the data was normalized by removing duplicates in multilabel attributes, converting similar values into a standardized form, and manually curating properties still present in their Wikidata-codified form.

The character coverage widely varies depending on the attribute, as shown in Table~\ref{gold_ds}. Specifically, the \textit{Age} attribute, which is critical to our task, initially had only 40 instances filled in. For this reason, we proceeded with manually annotating a subset of characters with their \textit{Age} value.

\begin{table}[!t]
\centering
\renewcommand{\arraystretch}{1}
\begin{tabular}{lcc}
\toprule
\textbf{Age category} & \textbf{F1-score} & \textbf{Instances} \\
\midrule
Child & 0.78 &  16 \\
Teenager & 0.44 & 10 \\
Adult & 0.90 & 242 \\
Senior & 0.71 &  25 \\
\bottomrule
\end{tabular}
\caption{Per-class F1-scores of human annotations for the \textit{Age} attribute. \textit{Adult}, the most represented class, shows a high level of agreement, while \textit{Teenager} shows only moderate agreement.}
\label{tab:age_f1}
\end{table}

\begin{table*}[!ht]
\centering
\small
\begin{tabular}{lcccl}
\toprule
\textbf{Attribute} & 
\textbf{Prediction} & 
\textbf{Gold reference} & 
\textbf{Human score} & 
\textbf{Interpretation Scale} \\
\midrule
Origin & America & France & 0.0 & Prediction is entirely incorrect \\
Occupation & Writer &  Author, nurse & 0.5 & Partially correct, lacks some occupations \\
Physical health & Speech disorder & Stuttering & 0.66 & Not precise but correct\\
Type & Kitty & Cat & 1.0 & Correct or synonym \\
\bottomrule
\end{tabular}
\caption{Examples of the interpretation scale used by annotators to evaluate each systems' predictions on open-class attributes.}
\label{tab:mhas_examples}
\end{table*}

\subsection{Age Annotation}
\label{sec:age_annot}


Age information within Wikidata is usually provided as a numerical value.
However, knowing the exact age of a character is not required to shape its voice, and only approximating an age range typically provides precise enough information.
We therefore convert the \textit{Age} attribute into a categorical attribute, consisting in four classes: Child, Teenager, Adult, Senior.
The 40 characters whose ages were available as numeric values were automatically mapped into the corresponding class.
Then, we manually annotated 319 additional \textit{Age} instances, resulting in a total of 359 annotated cases.
Age categories were inferred from websites that provide character analyses (e.g. GradeSaver\footnote{https://www.gradesaver.com/}, LitCharts\footnote{https://www.litcharts.com/})
We designed guidelines\footnote{Appendix \ref{app:age-guidelines}} that prioritize explicit mentions of age, followed by relative age cues, clear references (e.g., “child”), and stage of life indicators.
Uncertain cases were left blank. 

A total of 293 characters were double-annotated, from which we measured inter-annotator agreement with standard and soft variants of Cohen’s Kappa and F1-score.
Unlike other closed-class attributes, \textit{Age} is organized linearly with a natural progression from “Child” to “Senior”.
Therefore, discrepancies in adjacent classifications (e.g. "teenager" instead of "adult”) are less severe than distant ones (e.g. "child" instead of "senior") and could be due to the inherent difficulties in opting between two adjacent categories for certain characters.
For this reason, we also report a soft version of both metrics, assigning partial credit when the two annotations differed by only one adjacent age category, with a 0.8 weight.

We report a Cohen’s $\kappa$ coefficient of 0.56 in its standard form, increasing to $\kappa= 0.74$ when computing the soft variant, and also report a soft F1-score of 0.86.
These observations suggest a moderate agreement between annotators, highlighting that most disagreements occur between adjacent classes.
Table~\ref{tab:age_f1} displays per-class F1-scores for the 293 double-annotated instances, suggesting that "Teenager"  is the age category with most disagreement. 
Annotators finally performed an adjudication step to resolve remaining disagreements and refine the guidelines.
After this process, 359 characters have a filled \textit{Age} attribute.
Age categories distribution in the dataset are reported in Figure \ref{fig:distributions}.



\section{S-VoCAL Evaluation Framework}
\label{evaluation_framework}

We propose a dedicated evaluation framework tailored to each attribute type and data format (closed, semi-closed, and open). Closed and semi-closed attributes are evaluated using F1-based metrics, while open attributes are evaluated with a similarity metric based on Qwen-3 embeddings, similar to BERTScore, that we validate through correlation with human judgments.
These judgments are further used to map similarity values to a human-interpretable scale, allowing a better understanding of models' performances on open attributes.

\subsection{Closed and Semi-Closed Class Evaluation}

Closed-class attributes (\textit{Age}, \textit{Gender} and \textit{Type (human/non-human)}) take their values into a small set of candidates, thus standard classification metrics can be used normally.
Since most characters are of \textit{Type} "human" in S-VoCAL (91\%), we conduct a two-stage evaluation for this attribute.
The first stage assesses how accurate a system is at distinguishing between human and non-human characters, while the second accounts for semantic similarity between non-human characters.
\textit{Type (human/non-human)}, \textit{Age}, and \textit{Gender} are evaluated using Weighted F1 to account for label unbalance in each categories.

Similarly to the approach taken in the manual annotation of \textit{Age}, we propose a soft F1-score, which penalizes less errors in neighboring categories. The custom soft F1-score uses a weight matrix that encodes linear distances between age groups: 1.0 for exact matches, 0.8 for adjacent categories, and 0.0 otherwise.

The only semi-closed class attribute, \textit{Spoken languages}, takes its values into a broader set of candidates, yet limited to the set of languages names, which reduces the semantic ambiguity compared to open attributes. It can therefore be evaluated with a metric using exact match between values, such as F1-score. We evaluate \textit{Spoken Languages} as a multi-label setup, using binarization per class followed by micro-F1, which is suited to evaluate a system's ability to infer the set of (potentially multiple) languages spoken by each character. 

\subsection{Open-Class Evaluation}
\label{sec:qwenattrscore}

The open-class attributes (\textit{Origin}, \textit{Residence}, \textit{Occupation}, \textit{Physical Health}, \textit{Type (non-human)}) can be expressed in variable ways while bearing a very similar meaning, for instance, “England” vs. “United Kingdom” for \textit{Residence}, or “nobleman” vs. “count” for \textit{Occupation}.
BERTScore \cite{bertscore} has been extensively used in the context of the evaluation of generative systems, and could theoretically be applied in our evaluation framework.
However, the application of BERTScore within the context of S-VoCAL open-class attributes may entail the following potential limitations.
First our gold references and system predictions are typically lists of words or entities, or short phrases, depraved of contextual dependencies, limiting the expressivity of BERT-based representations \cite{ethayarajh-2019-contextual}.
A second issue is the reported inability for BERTScore to provide a faithful assessment of text quality in summarization \cite{fabbri2021summevalreevaluatingsummarizationevaluation}, which led the research community towards LLM-based automatic evaluation metrics \cite{min-etal-2023-factscore}.



\paragraph{Qwen3-based Metric} To address these limitations, we propose to evaluate semantic similarities between gold and predicted attributes using attribute-aware representations computed from Qwen3-8b Embeddings\footnote{https://huggingface.co/Qwen/Qwen3-Embedding-8B}, a recent instruction-aware embedding model that computes text representations conditioned on an instruction.
We leverage this sensitivity to instructions to derive small instructions prompts for each attribute, leading to attribute-conditioned representations of the gold references and predicted outputs.
Then, we simply compute the cosine similarity between the attribute-conditioned gold and predicted representations.




\begin{table}[!t]
\centering
\small
\setlength{\tabcolsep}{4pt}
\begin{tabular}{lcccc}
\toprule
\textbf{Attribute} & \textbf{$\alpha$} & $\rho$-Qwen3 & $\rho$-BERT & \textbf{n}\\
\midrule
Origin & 0.95 &  0.85 & 0.44 & 66\\
Residence & 0.93 & 0.85 & 0.31 {\scriptsize ($p=0.013$)} & 65\\
P. Health & 0.74 & 0.52 & -0.10 {\scriptsize($p=0.572$)} & 37\\
Occupation & 0.75 & 0.66 & 0.32 & 121\\
Type (n-h.) & 0.92 &  0.86 & 0.63 & 57\\
\bottomrule
\end{tabular}
\caption{Inter-annotator agreement ($\alpha$) and Spearman correlation ($\rho$) between human scores and automatic metrics. 
All correlations are significant at $p<0.001$, except where specified
\textit{Abbreviations:} P.~Health = \textit{Physical Health}; Type (n-h.) = \textit{Type (non-human)}.}
\label{tab:corr_human_eval}
\end{table}

\paragraph{Validation} We validate the proposed metric quality by computing Spearman $\rho$ correlations between the cosine similarity scores and human judgments.
To produce the judgments, three annotators evaluated a subset of the predicted open-class attributes on a discrete scale from 0 to 1, where 0 indicates an entirely incorrect prediction, 1 a correct prediction, and intermediate scores reflect partial correctness. To ensure consistent annotations, we provide to each annotator an interpretable judgment for each score, complemented with examples (some of these examples are displayed in Table~\ref{tab:mhas_examples}, and guidelines can be found in Appendix \ref{app:evaluation-guidelines}).

We report inter-annotator agreement as interval Krippendorff's $\alpha$ \cite{krippendorff2011computing} and the Spearman $\rho$ correlation between human judgments and both the Qwen3-based metric and BERTScore in Table~\ref{tab:corr_human_eval}.
Human annotations show a high level of agreement for all attributes, with Krippendorff's alpha ranging from 0.74 to 0.95.
Besides, the proposed metric shows a consistent and significantly higher correlation with human judgments than BERTScore on all attributes, with Spearman $\rho$ ranging from 0.52 to 0.85 while BERTScore achieves zero-correlations for Physical Health and a maximum value of $\rho=0.63$.
Interestingly, correlations between the Qwen-3 based metric and human judgments was found to be lower on the \textit{Physical Health} and \textit{Occupation} attributes that also display lower agreement between annotators, which might indicate that comparing gold and predicted values for these attributes is more challenging.


\begin{figure*}[!ht]
\centering
\includegraphics[width=.9\textwidth]{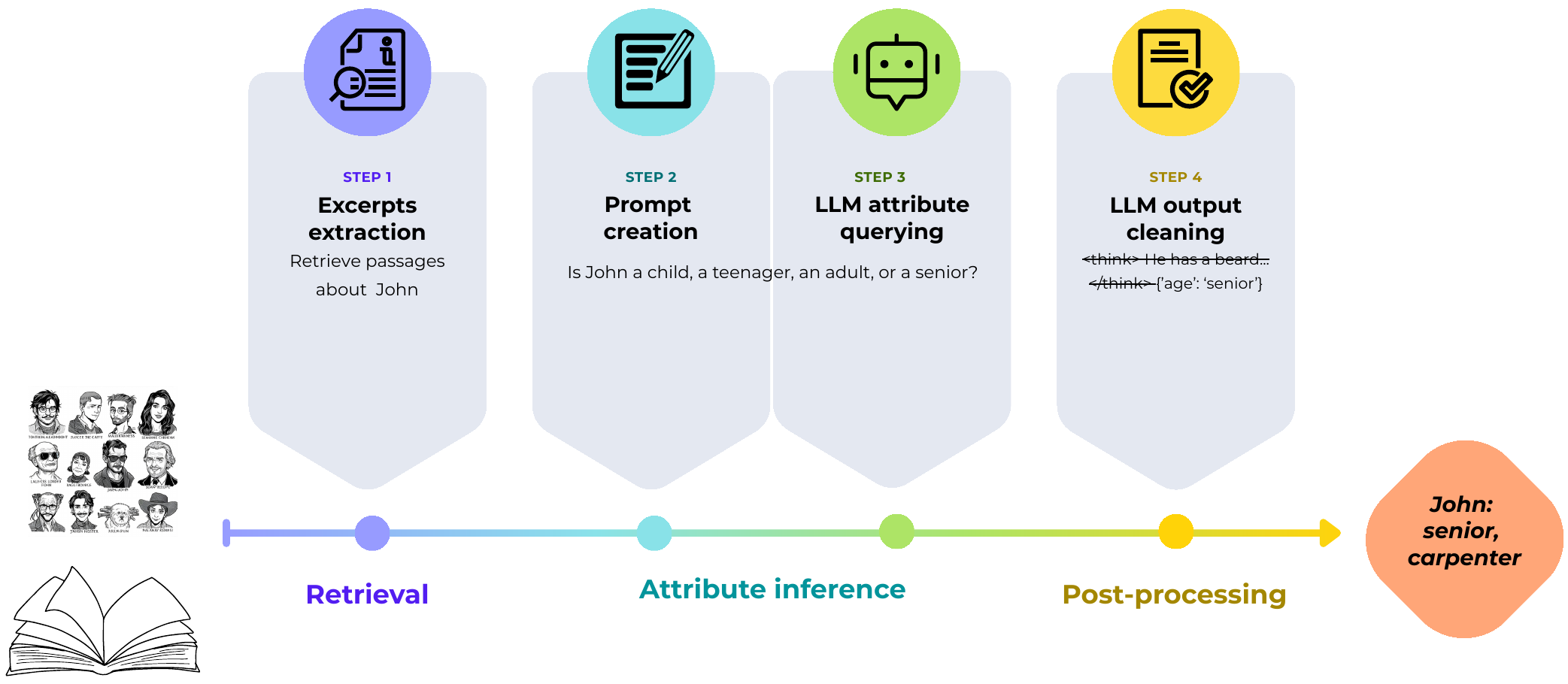}
\caption{Overview of the RAG pipeline. The first step extracts attribute-specific relevant excerpts for a character. These passages are then prompted to an LLM along with an attribute prompt.}
\label{fig:pipeline}
\end{figure*}

\paragraph{Human-Aligned Score} While the above correlation analysis suggests that our proposed metric is more reliable than BERTScore for evaluating open attributes, the raw cosine similarity scores lack direct human interpretability.
Thus, we further leverage human judgments and compute a Human-Aligned Score (HAS) by fitting an isotonic regression line to map the Qwen3 similarity values onto the human evaluation scale for each attribute.
This process automatically aligns raw similarity scores to interpretable values, allowing for a clearer understanding of model predictions.
Note that this approach only provides a partial interpretation of model predictions, subject to confounders in the isotonic regression.
However, we found empirically that HAS were typically well aligned with human judgments and easier to interpret than cosine similarity scores, and thus propose to use HAS in S-VoCAL's evaluation framework.

\section{Attribute Inference with RAG}

Following prior works \cite{papoudakis-etal-2024-bookworm} we perform Attribute Inference on S-VoCAL with Retrieval-Augmented Generation (RAG).
Given a novel and the character's identifiers (name, aliases or nicknames), the RAG pipeline derives character's attributes from relevant book excerpts.
RAG is particularly suited in the Attribute Inference scenario, where relevant information about a character's attribute might be disseminated in various book sections, or, in the worst-case scenario, might be implicitly mentioned once in the entire narrative.
By retrieving and processing attribute-relevant book passages, we address the particular problem of ``a needle in a haystack'' that occurs when processing an entire book at once.
The RAG pipeline is presented in Figure~\ref{fig:pipeline}, and is composed of three main components described in the following subsections.



\subsection{Retrieving Character Excerpts}
\label{retrieval}


The retrieval component selects the most relevant passages mentioning the character based on semantic similarity.
Using the character’s name and aliases, all named mentions in the novel were first located with regular expressions, and 200-word windows were then extracted around each mention.
We set this number to 200 experimentally, as it provides a good compromise between capturing relevant information and keeping the context size within the models’ limits.
We then compute passage embeddings with E5-large\footnote{https://huggingface.co/intfloat/multilingual-e5-large-instruct} \citep{wang2024textembeddingsweaklysupervisedcontrastive}, selected for its lightweight architecture and its query-based sentence embeddings, as well as its recent application in literary analysis, where it has proven effective in literary contexts \citep{hatzel-biemann-2024-story}.
To derive attribute-specific relevant excerpts, we perform the following steps:


\begin{itemize}
\item{\textbf{Instruction:}} We define a general instruction prompt, common to all attributes query

\item{\textbf{Attribute Query:}} For each target attribute, a specific query is created (e.g., for Age: “Is the character a child, a teenager, an adult, or a senior?”). All queries are available in Appendix \ref{app:attribute-queries}.
\item{\textbf{Scoring:}} We encode all queries and passages representations, then compute cosine similarities between both.
\item{\textbf{Filtering:}}
We select the ten most similar passages per attributes, and when multiple attributes are requested, duplicates across top-10 lists are removed. This list will be used as the input to the LLM-based Attribute Inference step.\end{itemize}



\begin{table}[!t]
\centering
\small
\renewcommand{\arraystretch}{1.25}
\setlength{\tabcolsep}{4pt}

\begin{tabular}{l l c c c}
\toprule
 & & Qwen3-8b & Phi4-14b & Baseline \\
\midrule
\multirowcell{3}{{\rotatebox[origin=c]{90}{\scriptsize \textbf{Weighted F1}}}}
    & Gender & 0.993 & \textbf{0.997} & 0.561 \\
    & Age & 0.783 & \textbf{0.800} & 0.660 \\
    & {Type {\scriptsize (h./n-h.)}} & 0.969 & \textbf{0.971} & 0.910 \\
\midrule
{\makecell{\rotatebox[origin=c]{90}{\scriptsize \textbf{Soft F1}}}}
    & Age & 0.927 & \textbf{0.941} & 0.908 \\
\midrule
\makecell{\rotatebox[origin=c]{90}{\scriptsize \textbf{Micro F1}}} 
    & \makecell[l]{{\small Spoken-}\\{\small Languages}} & 0.712 & \textbf{0.749} & 0.576 \\
    
\bottomrule
\end{tabular}
\caption{Results for close and semi-closed class attributes, with their dedicated metric. Type (h./n-h.) corresponds to the binary classification of Type Human vs Non-Human.}
\label{tab:merged_F1_results}
\end{table}

\subsection{Attribute Inference}
\label{attribute_inf}

Using the attribute-relevant excerpts derived from the retrieval step, we prompt an LLM to infer the requested character attributes.
A single prompt is constructed for each character, combining all requested attributes in one query. 
In this work, we use Qwen3-8B \citep{yang2025qwen3technicalreport}; and Phi-4 14B \cite{abdin2024phi4technicalreport} as LLM backbones.
Both models are quantized to 4 bits (Q4\_K\_M) to optimize memory usage and inference speed, 
and we kept default generation parameters.

We observed that the model outputs were often noisy and partially malformed (e.g., containing reasoning traces, Markdown artifacts, or broken JSON structures).
Thus, we apply a post-processing step to clean the predictions before converting them into structured character attributes (see Appendix \ref{app:postprocessing} for details).

\section{Results}

We report the results of the RAG-based pipeline on S-VoCAL, assessed through the proposed evaluation framework. 

\subsection{Closed and Semi-Closed Class Results}

For closed and semi closed class attributes, we compare LLM performance against a majority-value baseline, which always predicts the most frequent value for each attribute in S-VoCAL. The baseline provides a minimal reference point to contextualize the results obtained through the RAG-based pipeline.

We report results in Table~\ref{tab:merged_F1_results}.
For closed-class attributes, both RAG-based pipelines perform relatively well, and significantly better than the majority-value baseline. When using Phi-4 14B as the LLM backbone, we observe slightly higher scores than when using Qwen3-8B. Both systems obtain very high Weighted F1 scores for \textit{Gender} and \textit{Type}, while the results for\textit{ Age} are slightly lower.
We observe improved results when evaluating the Age attribute with soft-F1 which indicates that most errors occurred between adjacent classes, a behavior also reported in the manual annotations.



    

For the semi-closed attribute Spoken Languages, both RAG-based pipelines achieve considerable improvements over the baseline, with scores indicating a majority of correctly inferred languages.
However, the gap to the maximum score remains significant, suggesting occasional omissions or errors in the set of inferred spoken languages.



\begin{table}[!t]
\centering
\begin{tabular}{lcc}
\toprule
 & \multicolumn{2}{c}{\textbf{mHAS} $\uparrow$} \\ 
\textbf{Attribute} & Qwen3-8b & Phi4-14b \\
\midrule
Occupation      & \textbf{0.522} & 0.486 \\
Residence       & 0.498 & \textbf{0.502}   \\
Origin          & 0.420 & \textbf{0.422}  \\
Type (n-h.)& \textbf{0.785} & 0.781  \\
Physical health & 0.141 & \textbf{0.165}  \\
\bottomrule
\end{tabular}
\caption{Mean Human-Aligned Score (mHAS), for open attributes.
For most attributes, values of 0.5 indicates a partially correct prediction, with lacking information. Type (n-h.) = Type (non-human).}
\label{tab:isotonic_mhas}
\end{table}

\subsection{Open Class Results}

We report in Table~\ref{tab:isotonic_mhas} the mean Human-Aligned Scores (mHAS), derived from the similarity metric presented in Section~\ref{sec:qwenattrscore}.
We chose to report mHAS rather than raw similarity scores to facilitate interpretation, and refer the reader to Appendix \ref{app:similarity-scores} if interested in raw similarity scores.
mHAS ranges from 0 to 1, where a value of 0.5 often indicates a partial valid answer. We refer the reader to Appendix~\ref{app:evaluation-guidelines} for detailed interpretation values for each attribute.

\textit{Type} reaches the highest mHAS ($\approx$ 0.78), corresponding to mostly correct predictions, but lacking precision in some cases (e.g., predicting “bird” instead of “crow”).
\textit{Occupation} ($\approx$ 0.48--0.52) shows partially correct inference, with still some of the character's occupations missing from the prediction.
\textit{Residence} ($\approx$ 0.5) corresponds to predictions close to the expected value at a broader level, such as correct country but wrong region, or partially correct prediction, with part of the places to predict missing.
\textit{Origin} ($\approx$ 0.42) falls between high-level correctness (right continent but wrong country) and partial overlap (some correct places but also some omissions).
\textit{Physical Health} ($\approx$ 0.15) falls below the first interpretive level (0.33 = vague but correct distinction between poor and good health), implying that systems often fail to detect whether the character has a health condition or not.
These results suggest that while \textit{Type} is well retrieved, \textit{Occupation}, \textit{Residence} and \textit{Origin} are often partially correct, and \textit{Physical Health} remains challenging

Between the two LLMs pipelines, there is no clear winner for retrieving open-class attributes: we observe only minor fluctuations across attributes, except for \textit{Occupation} which appears slightly better inferred by Qwen3-8b. Interestingly, we observed no clear linear relationship between Qwen3 and Phi4 performance: if a model performs well on a specific instance, it does not necessarily imply that the other model also perform well on this instance, suggesting that both models have independent predictive behavior.
While the reasons behind this independence remains unclear, we hypothesize that the two models might exhibit different book memorization patterns \cite{chang-etal-2023-speak}, which could lead to a model performing well on instances of a particular book that the other model has not memorized, suggesting worse performance on this book.
We further detail this point in Appendix \ref{app:phi4_qwen3_distribution}.

\section{Conclusion}

This work introduces S-VoCAL, a dataset and evaluation framework designed to assess how well character attributes relevant to the design of a character’s voices can be inferred from novels.
S-VoCAL entails a broad range of attributes grounded in sociophonetic studies, and covers 192 books and 952 characters.
Given the attribute heterogeneity, we designed an extensive evaluation framework that includes standard metrics for closed and semi-closed cases, and a novel semantic similarity metric based on recent Large Language Models for open-class attributes.
S-VoCAL thus offers a standardized benchmark for evaluating systems designed to infer voice-relevant character attributes from novels.
We show the applicability of S-VoCAL by applying a simple RAG pipeline that infers voice-related attributes for a given character from entire novels.
Our results suggest that this pipeline handles closed-class attributes (e.g., \textit{Age}, \textit{Gender}, \textit{Type}) reliably, but remain limited in handling open attributes such as \textit{Physical Health} or \textit{Origin}. 
S-VoCAL contributes to the creation of immersive synthetic book narration, where each character is assigned a voice consistent with their identity.

\section{Limitations}

While this work presents an effort towards the creation of a dataset and evaluation framework for inferring voice-related attributes of fictional characters, it bears several limitations.

First, S-VoCAL only contains publicly available books drawn from Project Gutenberg.
These books often entails canonical literature, and are very likely memorized by most Large Language Models (LLMs).
This limitation is crucial when evaluating LLM-based attribute inference, as their reported performance on S-VoCAL does not necessarily imply that they will perform well on recent books, unseen during pre-training.
However, we believe that S-VoCAL structured evaluation framework remains useful when evaluating on such recent books.
Alternatively, book memorization and the extended knowledge of LLMs on books seen during pretraining -- which positively impact the accuracy of predicted attributes -- might play a crucial role in the faithful voice design of widely known characters.


Besides, our manual annotations only covered a subset of the \textit{Age} attribute, as we believe it was the most crucial attribute to design faithful character voices.
However, other attributes present in S-VoCAL remain largely underrepresented, and extending this annotation to these attributes could be very beneficial to improve coverage and reliability of the evaluation across all attribute types.

In addition, S-VoCAL provides 8 static character attributes derived from Wikidata.
However, many of the selected attributes are likely to evolve over time, such as Age, Physical Health or Spoken-Languages.
We found that recovering these dynamic patterns from any sources, either Wikidata or analysis websites, was particularly challenging, and decided instead to opt for static attributes that serve as a minimal standpoint to evaluate attribute inference systems.
However, we believe that the dynamic modeling of attributes is of crucial importance in the faithful representation of a character's voice throughout the story, and hope that S-VoCAL will serve as a first step towards this goal.

Finally, other character properties than those selected may have an influence on the character's voices. However, such characteristics are either highly contextual (e.g., their emotions) or subject to interpretative judgments of the reader, and are therefore less consistently represented in structured resources.
We believe that S-VoCAL provides a first step towards general character voice design, which can further be contextually rendered through additional inference layers that we leave as promising directions for future work.

\section{Ethical Considerations}

The dataset introduced in this work may carry structural biases, as Project Gutenberg, and especially books in English, mainly includes Western literary classics and Wikidata available information remains uneven across entities. Such imbalances can influence the types of characters and attributes that are most frequently represented and, consequently, evaluated. 

Moreover, future work on synthetic voice generation should ensure that the use of the defined attributes remains responsible, by creating clear ethical guidelines and monitoring practices, to avoid unintended stereotypical associations between character traits and speaking styles.

Finally, because S-VoCAL dataset was built exclusively from publicly available books, it remains imbalanced with a predominance of adult, male, and human characters. This distribution does not fully reflect the diversity of characters in literature, and including a more diverse set of books in future work could help address this limitation.

\section{Bibliographical References}\label{sec:reference}

\bibliographystyle{lrec2026-natbib}
\bibliography{lrec2026-example}

\bibliographystylelanguageresource{lrec2026-natbib}
\bibliographylanguageresource{languageresource}

\newpage
\appendix
\section*{Appendices}
\addcontentsline{toc}{section}{Appendices}

\section{Annotation Guidelines for Character Age}
\label{app:age-guidelines}

We describe below the guidelines given to human participants to annotate character age.
Characters are grouped into four age categories:

\begin{itemize}
    \item \textbf{Child}: From 0 to 12 year old
    \item \textbf{Teenage}: From 13 to 17 year old
    \item \textbf{Adult}: From 18 to 59 year old
    \item \textbf{Senior}: From 60 year old and beyond
\end{itemize}


\subsection{Sources}
Characters must be looked up online using their full name along with the book title, so their age can be inferred from web sources such as well-known websites Wikipedia, SparkNotes, GradeSaver, LitCharts, and others. This type of website usually includes plot summaries of the book and character descriptions or analyses, from which character’s age can be inferred. As the goal is to get trustworthy information, cross-sourcing the annotation in encouraged. If multiple sources provide conflicting information, choose the most consistent description or leave the annotation blank if uncertainty remains.

\subsection{General Principle}

The age category should be assigned using the following priority order:

\begin{enumerate}
    \item Explicit age stated in the text.
    \item Relative age information (e.g., ”younger than his 18-year-old sister”).
    \item Direct descriptive terms or keywords (e.g., ”boy”, ”old man”, ”teenage girl”).
    \item Behavior, role, or life situation (e.g., attending school, having a stable job, being married or a parent).
\end{enumerate}

\subsection{Steps of Decision for Annotation}

\begin{enumerate}
    \item \textbf{Is a numerical age explicitly stated?}\\
    $\rightarrow$ Assign the category using the table above.

    \item \textbf{If not, is the age given relative to another character?}\\
    Examples: "younger than his 18-year-old sister", a person described as older than someone who's already annotated as a senior.\\
    $\rightarrow$ Infer the most probable category.

    \item \textbf{If not, are there clear descriptive terms?}\\
    Examples: "child", "woman", "old man".\\
    $\rightarrow$ Assign the category accordingly.

    \item \textbf{If not, can behavior or social role indicate age?}
    \begin{itemize}
        \item Attending elementary school $\rightarrow$ Child.
        \item In high school $\rightarrow$ Teenager.
        \item Having a stable job or being married $\rightarrow$ Adult.
        \item Being described as "elderly" or "frail" $\rightarrow$ Senior.
    \end{itemize}

    \item \textbf{If no reasonable conclusion can be made:}\\
    Leave the annotation blank.
\end{enumerate}

\subsection*{4. Special Rules for Non-Human Entities}
\begin{itemize}
    \item If the age is stated and realistic (e.g., an elf "20 years old"), apply the same rules as for humans.
    \item If the age is unrealistic (e.g., a "200-year-old vampire"), assign the category based on the character's behavior and appearance.
    \item If no age is stated, apply the same decision steps as for humans.
\end{itemize}

\subsection*{5. Examples of Common Keywords}

We provide in Table~\ref{tab:app_keywords} a list of common keywords used by age-category.

\begin{table}
\centering
\setlength{\tabcolsep}{6pt}
\renewcommand{\arraystretch}{1.5}

\begin{tabular}{l|p{5.25cm}}
\hline
\textbf{Category} & \textbf{Frequent Keywords / Expressions} \\
\hline
Child & baby, infant, toddler, boy, girl, little, kid, schoolboy, schoolgirl \\
\hline
Teenager & teen, adolescent, high school student, schoolboy (high school), young man, young woman (could also be "adult" depending on the context)  \\
\hline
Adult & man, woman, husband, wife, worker, soldier, parent, teacher, merchant \\
\hline
Senior & old man, old woman, elderly, grandfather, grandmother, retired, frail, aged \\
\hline
\end{tabular}
\caption{Frequent keywords used to infer age category.}
\label{tab:app_keywords}
\end{table}

\section{Guidelines for Assessing Model Predictions}
\label{app:evaluation-guidelines}

We describe below the guidelines given to human participants to assess model predictions in open-class attributes.
These guidelines include examples per category that can be used to interpret our mean-Human-Aligned-Scores.

\subsection{General Evaluation Setup}

For each attribute, three elements are defined:
\begin{itemize}
    \item \{attribute\}\_pred: the model's prediction.
    \item \{attribute\}\_gold: the gold reference (correct answer).
    \item \{attribute\}\_agreement: the agreement score assigned by the annotator.
\end{itemize}

Only the agreement score is annotated. Scores are chosen from a predefined set of
values specific to each attribute.

\subsection{Origin and Residence}

Origin and Residence are evaluated as list attributes using the following
agreement scale:

\begin{itemize}
    \item \textbf{0.0}: Prediction is completely wrong (e.g., unrelated to the question, wrong continent).
    \item \textbf{0.25}: Prediction is partially related at a high level (e.g., correct continent but wrong country).
    \item \textbf{0.5}: Prediction is partially correct at a lower level (correct country, but wrong region or sub-area) OR partially correct but missing places.
    \item \textbf{0.75}: Prediction is correct but less specific than the gold, or overly specific and not fully verifiable.
    \item \textbf{1.0}: Prediction is fully correct and fully correspond to the gold, synonyms included. In the case of cities, accept cities which are very close to each other. If gold is a country, accept capital city or very big city in the country.
\end{itemize}

\paragraph{Examples}
\begin{itemize}
    \item 0.0: gold = ["France"], pred = ["America"]
    \item 0.5: gold = ["France", "Bretagne"], pred = ["France", "Marseille"]
    \item 1.0: gold = ["France"], pred = ["France", "Paris"]
\end{itemize}

\subsection{Physical Health}

\begin{itemize}
    \item \textbf{0.0}: Prediction not related or incorrect good/poor status.
    \item \textbf{0.33}: Vague prediction, but correctly identifies good/poor health.
    \item \textbf{0.66}: Correct but less specific than gold.
    \item \textbf{1.0}: Fully correct or synonymous prediction.
\end{itemize}

\paragraph{Examples}
\begin{itemize}
    \item 0.0: gold = ["Missing leg"], pred = ["good"]
    \item 0.33: gold = ["Cancer"], pred = ["ill"]
    \item 1.0: gold = ["Deaf"], pred = ["Deafness"]
\end{itemize}

\subsection{Occupation}

\begin{itemize}
    \item \textbf{0.0}: Completely wrong occupation.
    \item \textbf{0.25}: Same general domain but incorrect.
    \item \textbf{0.5}: Prediction partially correct but lacks some occupations not related at all to those found.
    \item \textbf{0.75}: Prediction in the same domain but not as precise as gold or prediction correct + unrelated occupation not mentioned in gold.
    \item \textbf{1.0}: Correct prediction or synonym, can be more precise than gold.
\end{itemize}

\paragraph{Examples}
\begin{itemize}
    \item 0.0: gold = ["writer"], pred = ["farmer"]
    \item 0.25: gold=["writer"] pred = ["editor"]
    \item 0.5: gold = ["novel author", "nurse", "mother"], pred = ["writer"]
    \item 1.0: gold = ["soldier", "captain"], pred = ["military sailor"]
\end{itemize}

\subsection{Type}

\begin{itemize}
    \item \textbf{0.0}: Completely wrong or human/non-human confusion.
    \item \textbf{0.33}: Wrong entity but correct human/non-human distinction.
    \item \textbf{0.66}: Prediction more general than gold.
    \item \textbf{1.0}: Correct prediction or synonym.
\end{itemize}

\paragraph{Examples}
\begin{itemize}
    \item 0.0: gold = ["human"], pred = ["cat"]
    \item 0.33: gold = ["cat"], pred = ["dog"]
    \item 0.66: gold = ["horse"], pred = ["equine"]
\end{itemize}

\section{Attribute Query Templates}
\label{app:attribute-queries}

For each character and each attribute, the following query templates are used for retrieval of the relevant passages with E5-large model:

\begin{itemize}
    \item \textbf{Age}: "Is \{character\_name\} a child, a teenager, an adult, or a senior?"
    \item \textbf{Gender}: "Is \{character\_name\} male or female?"
    \item \textbf{Origin}: "Where is \{character\_name\} from?"
    \item \textbf{Residence}: "Where does \{character\_name\} live?"
    \item \textbf{Spoken Languages}: "What languages does \{character\_name\} speak?"
    \item \textbf{Type}: "What type of entity is \{character\_name\}?"
    \item \textbf{Occupation}: "What is \{character\_name\}'s occupation?"
    \item \textbf{Physical Health}: "How is \{character\_name\}'s health condition?"
\end{itemize}

\begin{table}[b!]
\centering
\begin{tabular}{lccc}
\toprule
\textbf{Attribute} & \textbf{Qwen3} & \textbf{Phi4} & \textbf{Baseline} \\
\midrule
    Origin & 0.481 & 0.446  & \textbf{0.512}         \\
    Residence & 0.426 & \textbf{0.476} &   0.366  \\
    Type (non-human) & \textbf{0.869} & 0.810 &    0.569  \\  
    Occupation  & \textbf{0.561} & 0.558 &   0.403  \\
    Physical Health & 0.301 & \textbf{0.329} &   0.317    \\
    
\bottomrule
\end{tabular}
\caption{Mean cosine similarity scores per open attributes}
\label{tab:cos_results}
\end{table}

\begin{table*}[t]
\centering
\begin{tabular}{p{5cm} p{5cm} p{5cm}}
\toprule
\textbf{Input problem (example)} & \textbf{Cleaning operation} & \textbf{Expected result} \\
\midrule
Reasoning traces \texttt{<think>...</think>} & Remove reasoning segment & Traces removed \\
\hline
Markdown fences \texttt{```json ... ```} & Remove formatting markers & Keep only JSON \\
\hline
Invalid values \texttt{: None} & Normalize invalid value & Valid JSON \\
\hline
Multiple JSON blocks & Select final JSON block & Single target block \\
\hline
Unquoted list items \texttt{[Paris, London]} & Add missing quotes & \texttt{["Paris","London"]} \\
\hline
Extra brackets or commas \texttt{]],} & Repair unbalanced separators & Balanced brackets \\
\hline
Missing commas between objects & Repair list separators & Properly separated elements \\
\hline
Nested malformed lists \texttt{["["Paris""]} & Normalize list structure & \texttt{["Paris"]} \\
\hline
Literal \texttt{'null'} inside lists & Remove placeholder value & Cleaned list \\
\hline
\texttt{origin} as key--value pairs ("origin": ’country’: "France", "city": "Paris") & Flatten into list of values & \texttt{"origin": ["France","Paris"]} \\
\hline
Non-string input when string expected & Discard attribute & \textit{null} as safe default \\
\bottomrule
\end{tabular}
\caption{Cleaning rules applied to model outputs}
\label{tab:cleaning_rules}
\end{table*}

\section{Post-processing of Model Outputs}
\label{app:postprocessing}

We summarize in Table~\ref{tab:cleaning_rules}  the main post-processing operations applied to raw LLM outputs. Code for post-processing is available on the associated repository at \url{https://github.com/AbigailBerthe/S-VoCAL}.

\subsection{Example}
We present below some examples of the post-processing operation applied to the LLM outputs

\paragraph{Before post-processing (simplified):}
{\small
\begin{verbatim}
<think> (...) </think>
{
  "origin": ["Europe", Germany, "Parchim"],
  "residence": ["Germany", "Hamburg",],
  "spoken_languages": [
        "German", English, 'null']
}
\end{verbatim}
}

\paragraph{After post-processing:}

{\small
\begin{verbatim}
{
  "origin": [
        "Europe", "Germany", "Parchim"],
  "residence": ["Germany", "Hamburg"],
  "spoken_languages": [
        "German", "English"]
}
\end{verbatim}
}

\section{Raw Similarity Scores}
\label{app:similarity-scores}

Table~\ref{tab:cos_results} reports the mean cosine similarity scores for each open attribute.

\section{Distribution of Phi-4 and Qwen3 mHAS}
\label{app:phi4_qwen3_distribution}

\begin{figure}[h]
    \centering
    \includegraphics[width=1\linewidth]{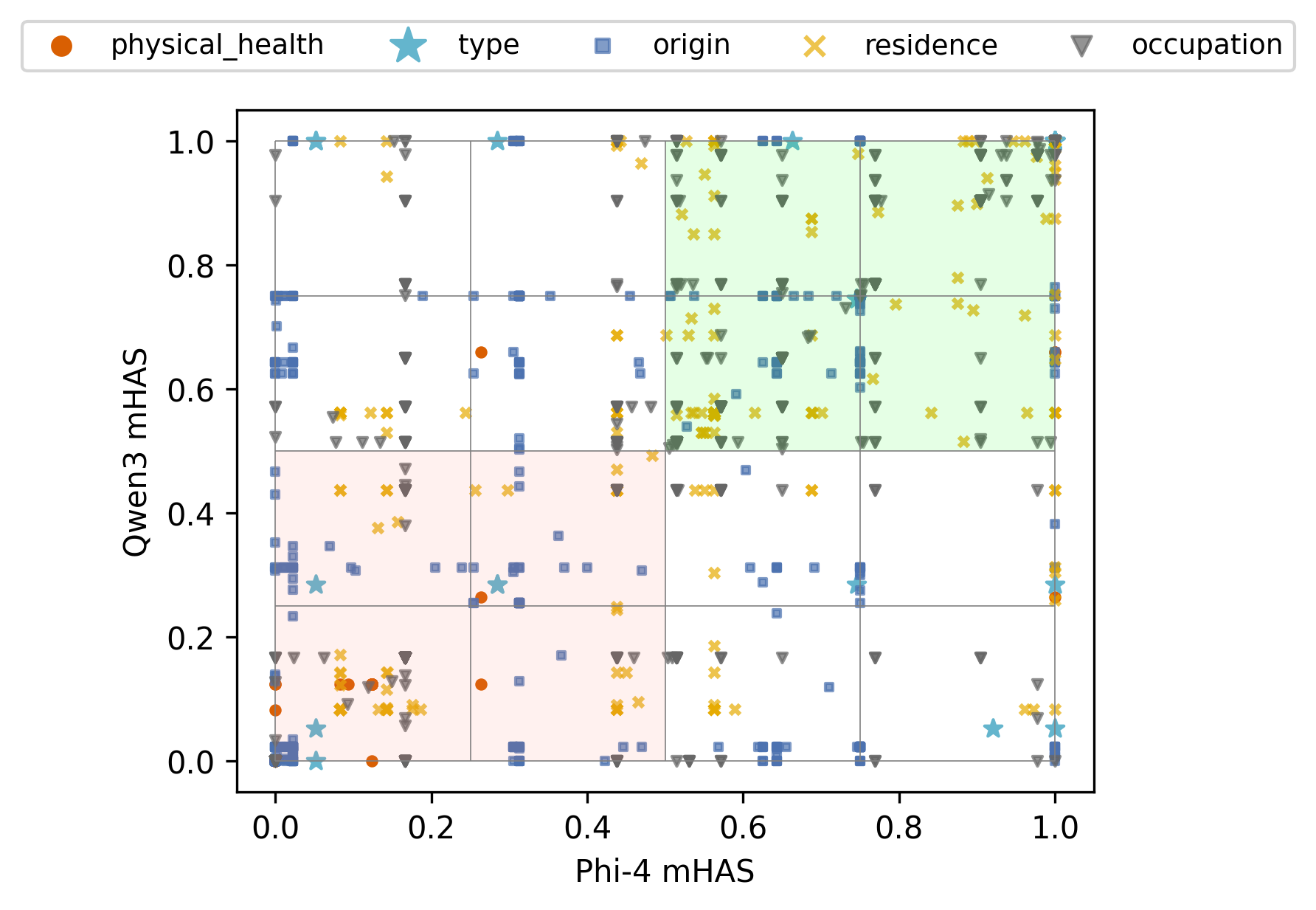}
    \caption{Comparison of Phi-4 and Qwen3 mHAS scores across attributes.}
    \label{fig:comp_phi4_qwen3}
\end{figure}

Figure~\ref{fig:comp_phi4_qwen3} presents an instance-level comparison of the mHAS scores assigned to outputs produced by Phi-4 and Qwen3, across the open-class attributes.
We see no linear relationship between model's performance, indicating that models tend to make errors on different instances, although provided with the same input.
We hypothesize that this behavior might be due to different book-memorization per model. 

\end{document}